\documentclass[runningheads]{llncs}
\usepackage{graphicx}
\usepackage{amsmath,amssymb} 
\usepackage{color}

\usepackage[inline]{enumitem}
\usepackage[utf8]{inputenc}
\usepackage{subcaption}
\usepackage{booktabs}
\usepackage{multirow}
\usepackage{adjustbox}
\usepackage{xspace}
\usepackage[table]{xcolor}
\usepackage{colortbl}
\usepackage[ruled,linesnumbered]{algorithm2e}
\usepackage[colorlinks,bookmarks=false,citecolor=cyan]{hyperref}

\newcommand{\latinphrase}[1]{\textit{#1}}
\newcommand{\etal}{\latinphrase{et~al.}\xspace}

\newcommand{\eg}{\latinphrase{e.g.}\xspace}

\newcommand{\acron}{FasTCo\xspace}

\newcommand{\video}{V}
\newcommand{\frm}{I}
\newcommand{\bbox}{b}
\newcommand{\model}{\Phi}
\newcommand{\ftgen}{\phi}
\newcommand{\cls}{C}
\newcommand{\prob}{p}
\newcommand{\logit}{q}
\newcommand{\lbl}{y}
\newcommand{\feat}{x}

\newcommand{\gaussian}{\mathcal{N}}
\newcommand{\bs}{m}
\newcommand{\std}{\delta}

\begin{document}

\pagestyle{headings}
\mainmatter
\def\ECCVSubNumber{243}  

\title{On Improving Temporal Consistency for Online Face Liveness Detection System} 


\titlerunning{On Improving Temporal Consistency for Online Face Liveness Detection}
\author{Xiang Xu \and Yuanjun Xiong \and Wei Xia}
\authorrunning{X. Xu et al.}
%
\institute{AWS AI, Amazon}
\maketitle

\begin{abstract}
In this paper, we focus on improving the online face liveness detection system to enhance the security of the downstream face recognition system.
Most of the existing frame-based methods are suffering from the prediction inconsistency across time. To address the issue, a simple yet effective solution based on temporal consistency is proposed. Specifically, in the training stage, to integrate the temporal consistency constraint, a temporal self-supervision loss and a class consistency loss are proposed in addition to the softmax cross-entropy loss. 
In the deployment stage, a training-free non-parametric uncertainty estimation module is developed to smooth the predictions adaptively.
Beyond the common evaluation approach, a video segment-based evaluation is proposed to accommodate more practical scenarios.
Extensive experiments demonstrated that our solution is more robust against several presentation attacks in various scenarios, and significantly outperformed the state-of-the-art on multiple public datasets by at least 40\% in terms of ACER. Besides, with much less computational complexity (33\% fewer FLOPs), it provides great potential for low-latency online applications.  
\end{abstract}

\section{Introduction}
\label{Sec:Intro}
Serving as a shield that enhances face recognition security, liveness detection aims at predicting whether the input sample is a real authenticated subject or a presentation attack attempt \cite{Galbally_2015_190908}.
To detect the presentation attacks, 
different models have been developed using spatial and temporal \mbox{information \cite{Yang_2019_190813}}, domain generalization \cite{Shao_2019_190908,Shao_2020_200225}, and zero-shot learning \cite{Liu_2019_190813}.

\begin{figure}[t]
\begin{center}
    \begin{subfigure}{.6\linewidth}
    \centering
    \includegraphics[width=\linewidth]{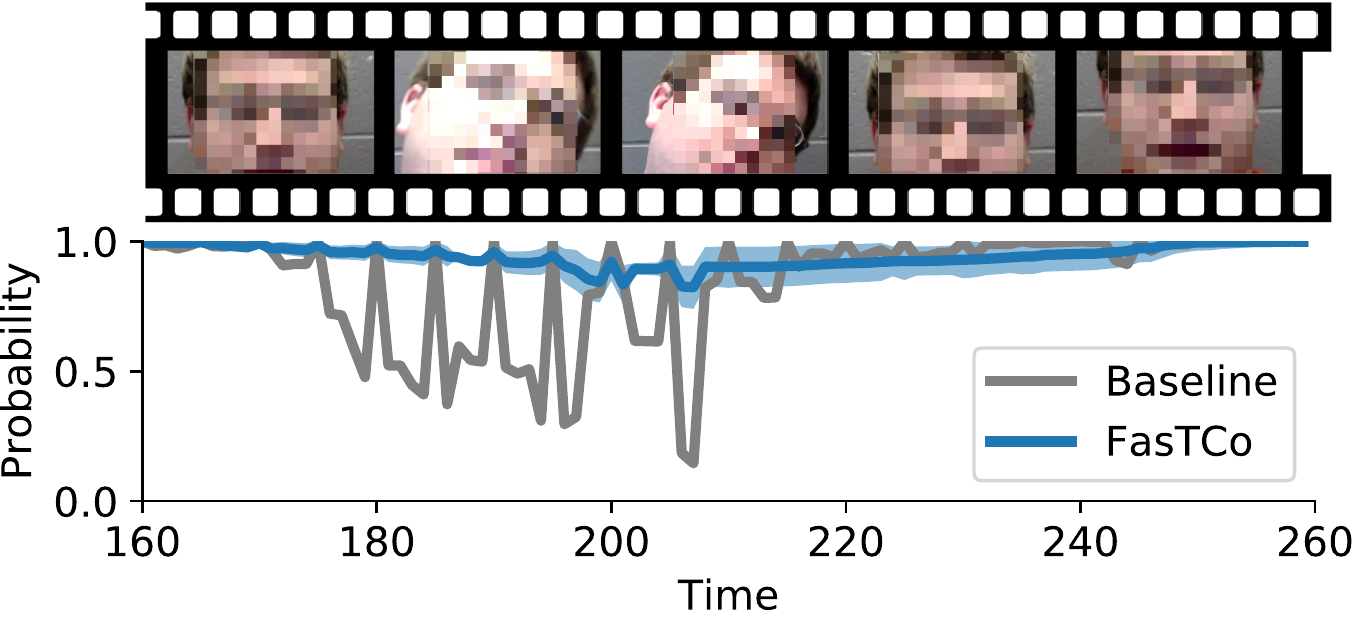}
    \caption{}
    \label{Fig:Intro-TC}
    \end{subfigure}
    \hfill
    \begin{subfigure}{.34\linewidth}
    \centering
    \includegraphics[width=\linewidth]{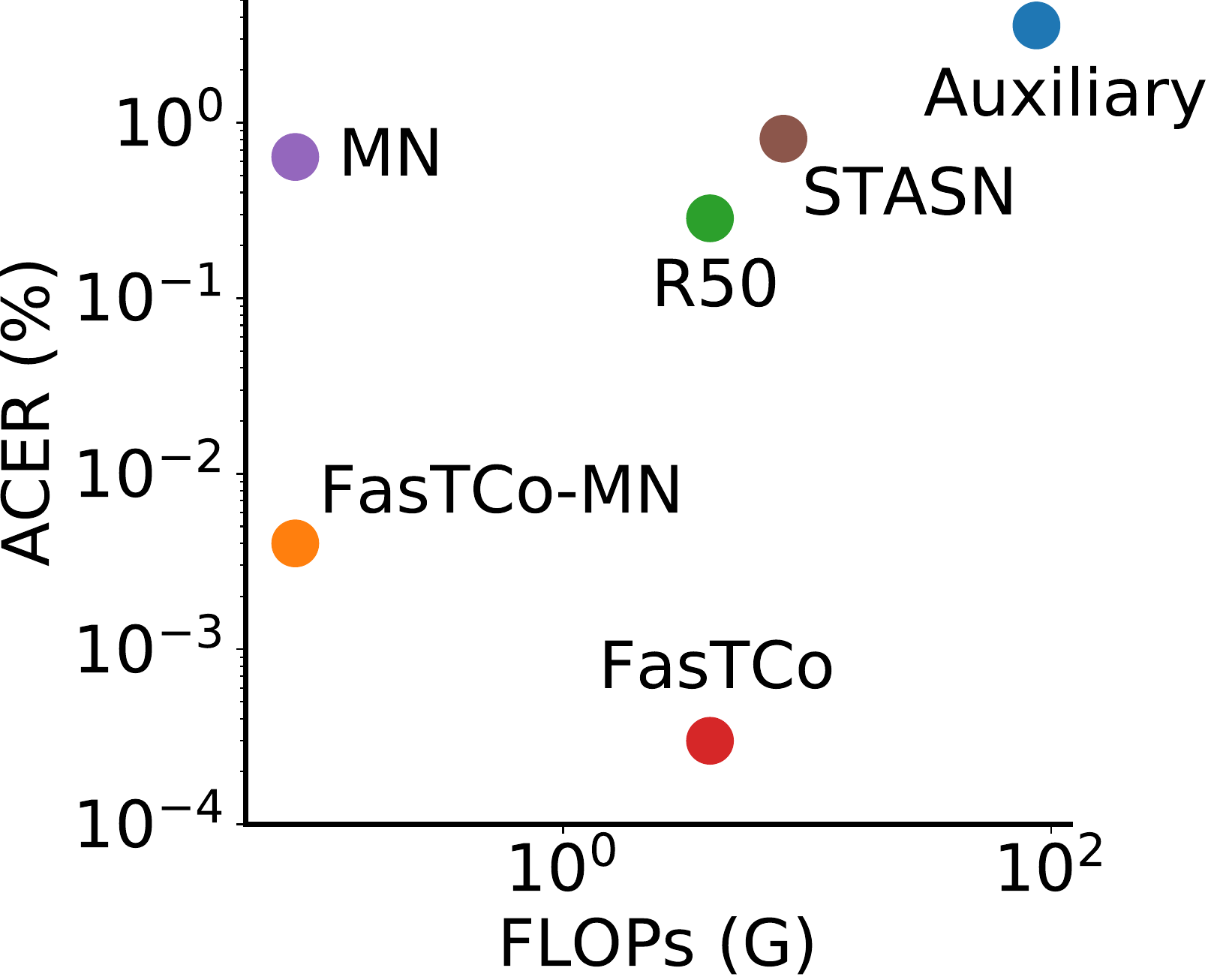}
    \caption{}
    \label{Fig:Intro-FLOP-ACER}
    \end{subfigure}
\end{center}
\vspace{-1.5em}
\caption{Depiction of (a) temporal inconsistency of the predictions on one video clip from a fame-based baseline model (gray curve) and the predictions with uncertainty estimations from our model - FasTCo (the blue curve represents prediction probability and the shade represents confidence levels, the face is blurred to hide identity); (b) model comparison of previous methods and two variations of \acron using ACER (\%) on SiW with protocol one and FLOPs (G). Bottom left is the best (Best view in color).}
\vspace{-1.5em}
\end{figure}

In this paper, we focus on online face liveness detection for common real-world use cases such as face authorization.
Unlike offline video analysis \cite{Wang_2018_190910} which can observe an entire video to make a final prediction, online processing requires low-latency prediction for each incoming frame.
In this setting, the most common approach is to predict the liveness probability per frame \cite{Liu_2018_190806,Liu_2019_190813}. 
However, as depicted in Fig.~\ref{Fig:Intro-TC}, such a frame-based model has larger prediction variance within a short period (the standard deviation of predictions is 0.2). 
By further analyzing the false positives and false negatives (presented in the supplementary material), we noticed that it tends to make unstable predictions when the subject undergoes large motion or illumination changes. 
Therefore, we hypothesize that one common underlying issue for the frame-based liveness detection systems is temporal inconsistency. 

To address this issue, a simple yet effective Face anti-spoofing system using Temporal Consistency (\acron) is proposed with temporal-aware model training and adaptive model predictions.
Specifically, in the training stage, beyond the softmax cross-entropy loss for multi-class classification, to enforce consistency of video sequences in the embedding space, two additional loss functions are proposed to improve the training, aiming to minimize the intra-class embedding distances for video sequences and presentation attacks, respectively. 
In the deployment stage, based on temporal consistency, a training-free uncertainty estimation module is developed to adaptively update the liveness probability, which results in much more consistent predictions. 
For instance, as depicted in Fig.~\ref{Fig:Intro-TC}, the liveness probabilities predicted by \acron on the same video clip have a much lower variation (standard deviation is 0.04) compared with the baseline model. 
This approach can be considered as a special way to select informative past frames in the online setting.
Additionally, it is a generic approach that can deploy a more lightweight backbone (\acron-MN) and still achieve better performance than state-of-the-arts, as shown in Fig.~\ref{Fig:Intro-FLOP-ACER}. Such lightweight model provides great potential for low-latency applications, especially on edge deployment environment such as mobile phone or IoT devices.

To evaluate the online models, besides the commonly used frame-based evaluation, a video segment evaluation approach is introduced to provide metrics for different application scenarios. 
Extensive experiments including the ablation studies were conducted, and have demonstrated that our method significantly outperforms the state-of-the-art on several publicly available datasets with at least 33\% fewer FLOPs.
On the SiW \cite{Liu_2018_190806} dataset, \acron obtained an ACER of $3 \times 10^{-6}$,  almost $0.1\%$ of state-of-the-art under the protocol one, while achieving at least 50\% relative improvement using other protocols.
Meanwhile, the proposed solution exceeds the state-of-the-arts by 40\%+ on \mbox{OULU-NPU \cite{Boulkenafet_2017_190908}}, \mbox{SiW-M \cite{Liu_2019_190813}}, and cross-domain datasets. 
In summary, our method is more robust against multiple factors in practical use cases such as unseen presentation attacks, illumination change, and acquisition devices.

With the temporal consistency, the following benefits can be expected:
\begin{enumerate*}[label=(\roman*)]
    \item Simple thresholding: because the prediction of the model is unstable in adjacent continuous frames, it is difficult to determine an appropriate threshold for liveness classification. The system would either have less security if the threshold is too low (APCER $\uparrow$) or bad user experience from false rejects when the threshold is too high (BPCER $\uparrow$). However, with temporal consistency, the system outputs a more consistent prediction, leading to a much easier balance of APCER and BPCER in real applications.
    \item Uncertainty estimation: with the proposed uncertainty module, in addition to the liveness score, the system outputs the uncertainty estimation, which can be used to filter out frames with highly uncertain predictions. This greatly enhances the robustness of the system.
\end{enumerate*}
In summary, the contributions of this paper are:
\begin{enumerate}[label=(\roman*), topsep=0.3em, itemsep=0em, parsep=0em]
    \item Temporal inconsistency was identified as a common issue of the current face liveness detection systems;
    \item A simple yet effective solution, including two additional losses and a training-free uncertainty estimation module, was proposed to significantly improve the model performance without extra complexity and latency;
    \item In addition to the common frame-based evaluation approach, a video segment-based evaluation was proposed to measure both the latency and accuracy of the model for different application scenarios. 
\end{enumerate}

\section{Related Work}
\label{Sec:LR}

The common presentation attacks \cite{Galbally_2015_190908} to face recognition systems include using print photos, video replay, and 3D masks. 
The recent face liveness detection methods to identify these attacks can be classified into two major streams in general:
\begin{enumerate*}[label=(\roman*)]
    \item Static approaches: Some image clues from color space and frequency domain \cite{Boulkenafet_2016_190910,Jourabloo_2018_190910} were used to detect artifacts. In addition, some human-crafted features such as LBP \cite{Boulkenafet_2016_190910} and the features learned by CNN \cite{Atoum_2017_190910,Mehta_2019_190910,Zhang_2019_190806,Shao_2019_190908} were extracted to train a binary classifier. Domain generalization and meta-learning techniques \cite{Xu_2019_190315,Sarafianos_2019_190910,Shao_2020_200225} have also been used to learn generalized feature representations \cite{Li_2018_190909,Li_2018_190910,Shao_2019_190908,Wang_2020_200430,Jia_2020_200509} from multiple domains to improve the generalization of the model. Liu \etal \cite{Liu_2019_190813} developed a deep tree-structured learning process to learn homogeneous features of presentation attacks in the upper nodes of the tree and distinct features to classify each specific attacks in the leaf nodes. 
However, such static methods do not consider the relationship across the temporal dimension, and thus lacking the temporal consistency in predictions. 
    \item Dynamic methods: The motion of the face, either part or as a whole, was used to predict the liveness. 
    Multiple features extracted from video frames were aggregated and the predictions were fused by Siddiqui \etal \cite{Siddiqui_2017_191114} to generate a liveness score. 
    Similar to the common approach in action recognition \cite{Wang_2018_190910,Shou_2018_191105,Xu_2019_191030,Gao_2019_191030}, both spatial and temporal information \cite{Xu_2015_190910,Yang_2019_190813} of a video clip have been explored to make the final decision based on a CNN-LSTM network \cite{Donahue_2015_200303}. Nevertheless, it is hard to learn the temporal information by jointly learning CNN and LSTM networks.
\end{enumerate*}
In this work, we improved the model training by introducing new loss functions and inference consistency with an uncertainty estimation module.

\section{Temporal Inconsistency}
\label{Sec:Motivation}

Compared with the sequential models \cite{Yang_2019_190813}, the frame-based model is easier to implement and can be directly used to process online untrimmed videos. 
To find the root cause of why the model fails in some cases, the ResNet-50, a commonly used network structure serving as the baseline in the literature \cite{Yang_2019_190813,Zhang_2019_190806}, was used to train a frame-based online liveness detection model with a softmax loss on SiW \cite{Liu_2018_190806} dataset.
The evaluation was performed following the protocol one of the dataset.
The videos were ranked according to the number of predictions errors. 
After analyzing the errors (see supplementary material), there are two observations: 
\begin{enumerate*}[label=(\roman*)]
    \item Time-wise, one common pattern across the false positive and false negative samples is the prediction inconsistency.
The predictions are not stable and there are many spikes in the estimated probabilities.
By analyzing the frames corresponding to the false predictions, we made a hypothesis that such sudden prediction change was due to the movement of the subject or the environment changes such as reflection.
    \item Prediction-wise, the probability outputs are either extremely high for false positives or low for false negatives at some time stamps, indicating that the model is over-confident of its predictions on outliers.
\end{enumerate*}
The top two videos (see the supplementary material) containing the highest false positive rate and false negative rate were selected for illustration.
Based on these observations, to improve the baseline model, we need to answer the following questions: 
\begin{enumerate*}[label=(\roman*)]
    \item How to use temporal information to improve the training of a single network model?
    \item How to use temporal consistency to increase the robustness of the model inference?
\end{enumerate*}
Therefore, this paper will focus on solving the temporal inconsistency issue from both training and inference stages. 

\begin{figure}[t]
\begin{center}
   \begin{subfigure}{.48\linewidth}
        \includegraphics[width=\linewidth]{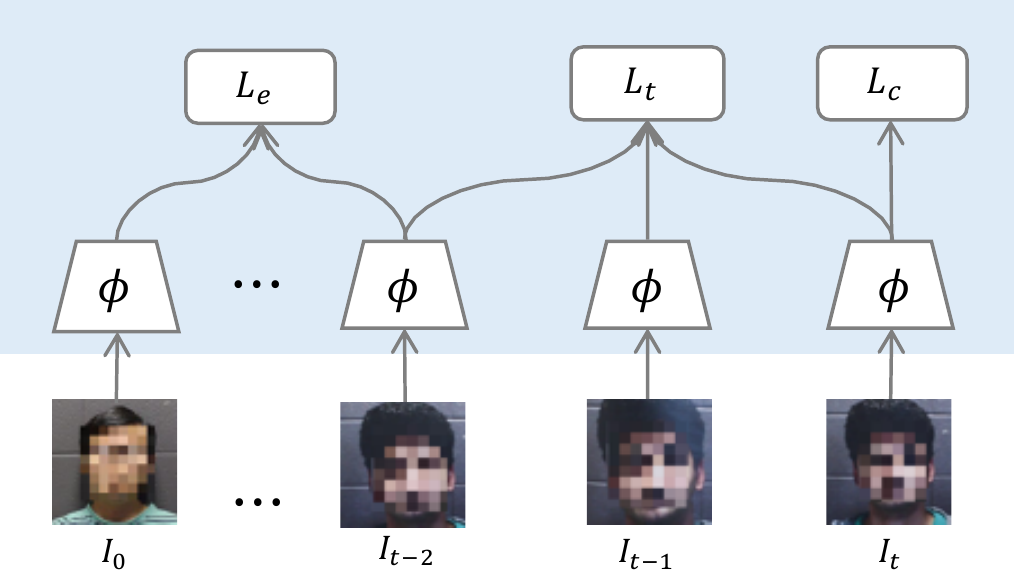}
        \vspace{-1.5em}
        \caption{}
        \label{Fig:Train}
    \end{subfigure}
    \ 
    \begin{subfigure}{.495\linewidth}
        \includegraphics[width=\linewidth]{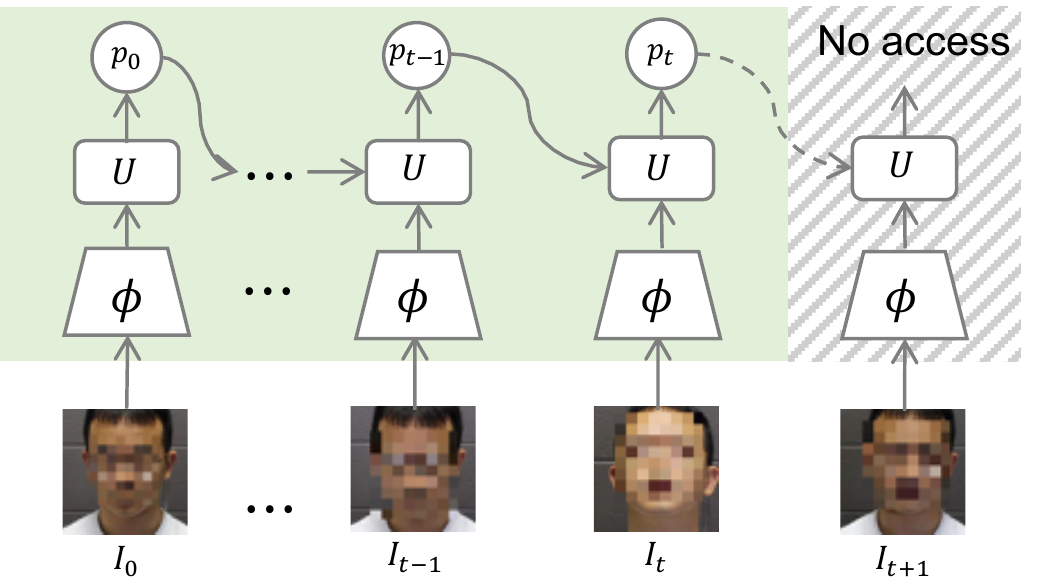}
        \vspace{-1.5em}
        \caption{}
        \label{Fig:Test}
    \end{subfigure}
\end{center}
\vspace{-1em}
\caption{Overview of (a) the training stage: in addition to the multi-class classification loss $L_c$, we propose a temporal self-supervision ($L_t$) loss on the features extracted from the same video sequence and a class consistency loss ($L_e$) to enforce the intra-class distances; (b) the deployment stage: a training-free uncertainty module $U$ is proposed to estimate the uncertainty based on the temporal consistency to smooth the liveness probabilities in the online setting.}
\label{Fig:Method}
\vspace{-1em}
\end{figure}

\section{Online Face Liveness Detection System}
\label{Sec:Method}
We first define two key properties for the online face liveness detection system, and then propose two strategies to improve the robustness of the model based on temporal consistency.

\subsection{Formulation}
Mathematically, a live video can be represented as a sequence of frames $\video = \{\frm_0, \ldots, \frm_t, \ldots, \frm_T\}$, where $t$ is the current time stamp and $T$ is the total number of frames.
To ensure the input sequence of faces belongs to the same identity, a face tracker \cite{Bergmann_2019_200227} can be deployed instead of a naive face detector to provide the temporal-spatial information for a sequence of face bounding boxes $\bbox_t$. 

\vspace{.5em}
\noindent \textbf{Input:} The input of the system is a cropped face from the video frame $\frm_i$ using its bounding box $\bbox_i$ from tracking. From now on, we will use $\frm_i$ to represent the face region frame for simplicity.

\noindent \textbf{Output:}
The output of the system is a liveness probability $\prob_t$, where a binary decision $\lbl$ (live or attack) can be determined with a threshold.

\noindent \textbf{Model:}
Usually, a face liveness detection model $\model(\cdot)$ consists of a feature extractor $\ftgen(\cdot)$ and a classifier $\cls(\cdot)$.
The liveness logit $\logit_t$ can be obtained by forwarding the current face frame $\frm_t$ to the network denoted as $\logit_{t} = \cls(\ftgen(\frm_t)) = \cls(\feat_t)$, where $\feat_t$ is the feature representations of the frame $\frm_t$.
The liveness probability $\prob_t$ can be generated by applying a normalization activation function such as softmax or sigmoid to the logit $\logit_{t}$.
As an additional constraint due to the online setting, when making a prediction on the frame $\frm_t$, the model $\model(\cdot)$ can only use the information from $[\frm_0, \frm_t]$ but is forbidden to access $[\frm_{t+1}, \frm_T]$ (Fig.~\ref{Fig:Test}).

\subsection{Temporal Consistency Properties}
By deploying the face tracker, the temporal consistency comes with the following two properties:
\begin{property}[Identity Consistency]
There is only one subject identity in the input stream. 
\end{property}
\begin{property}[Prediction Consistency]
The model should have consistent predictions on the frames within the same video tracklet.
\end{property}

\subsection{Improving Consistency}
Applying the properties of temporal consistency, the following loss functions are used to train the network in an end-to-end manner (Fig.~\ref{Fig:Train}) and a uncertainty module is proposed to keep prediction consistency in the deployment stage (Fig.~\ref{Fig:Test}):

\noindent \textbf{Classification Supervision}: 
Unlike the previous methods \cite{Liu_2018_190806,Jourabloo_2018_190910,Yang_2019_190813,Liu_2019_190813} that trained a binary classifier, the labels of video types (\eg, print, video replay) can be used as the supervision to train a multi-class classifier using a softmax cross-entropy loss: 
\begin{equation}
L_c = - \frac{1}{\bs} \sum_{i=0}^{\bs} \log \prob_{\lbl_i},
\end{equation}
where $\bs$ is the batch size.
The benefits of converting a binary classification problem into a multi-class classification setting are in three-folds:
\begin{enumerate*}[label=(\roman*)]
    \item The discriminative features to distinguish different types of presentation attacks can be learned;
    \item The embedding space of liveness class can be squeezed into a more compact space than using a binary classification, which helps decrease the false positive rate;
    \item Fine-grained analysis can be conducted when the model makes mistakes. 
\end{enumerate*}

\noindent \textbf{Temporal Consistency Self-Supervision}: 
To keep temporal consistency across multiple frames (\eg, frames within the same video tracklet), a self-supervision loss is proposed to regularize the intra-video consistency in the embedding space, denoted as:
\begin{equation}
L_{t} = \frac{1}{\bs} \sum_{i=0}^{\bs} \max_{i,j \in v} ||\feat_i - \feat_j||_2^2,
\end{equation}
where $\feat_i$ and $\feat_j$ are the feature representations of two frames from the same video clip $v$ within the batch.

\noindent \textbf{Class Consistency Loss}: 
Similar to the temporal consistency self-supervision, the embedding learned from the same class but from different videos should be as close as possible. Therefore, a class consistency loss function can be defined as follows:
\begin{equation}
L_{e} = \frac{1}{\bs} \sum_{i=0}^{\bs} \max \lbl_{ij}||\feat_i - \feat_j||_2^2,
\end{equation}
where $\lbl_{ij}$ is equal to 1 when $\feat_i$ and $\feat_j$ belong to the same class within the batch, otherwise $\lbl_{ij}$ is 0.
In the end, the final loss can be formulated as:
\begin{equation}
L = L_{c} + \beta L_t + \gamma L_e.
\end{equation}

\subsubsection{Filtering with Uncertainty Estimation}
During the deployment stage, to keep the prediction consistent within the same tracklet, a simple yet effective solution is proposed to estimate the model uncertainty and smooth the model predictions adaptively.

Due to the online setting, the uncertainty module can only observe the historical logit outputs $\{\logit_0, \ldots, \logit_t\}$ from the model.
Based on the temporal consistency, we can assume that: 
\begin{enumerate*}[label=(\roman*)]
    \item The random variable of the liveness score $\pi_t$ at the time step $t$ follows a Gaussian distribution $\gaussian(\hat{\mu}_t, \hat{\std}_t^2)$, where $\hat{\mu}_t$ and $\hat{\std}_t$ denote the moving average and standard deviation of $\pi$;
    \item The single logit observation $\logit_t$ follows another Gaussian distribution $\gaussian(\pi_t, \std^2)$.
\end{enumerate*}
According to the Bayesian rule, the posterior can be written:
\begin{equation}
\label{Equ:Loglikelihood}
\begin{split}
& \prob(\pi_{t}|\pi_{0}, \ldots, \pi_{t-1}; \logit_{t}) = \frac{\prob(\pi_{0}, \ldots, \pi_{t-1}; \logit_{t} | \pi_{t})\cdot \prob(\pi_{t})}{\prob(\pi_{0}, \ldots, \pi_{t-1}; \logit_{t})} \\
&= \alpha \cdot \prob(\logit_{t} | \pi_{t}) \cdot \prob(\pi_{t}| \pi_{0}, \ldots, \pi_{t-1}),
\end{split}
\end{equation}
where $\alpha$ is a normalization constant.
Based on the assumptions of temporal consistency, the Equ.~\eqref{Equ:Loglikelihood} can be derived into the following equation:
\begin{equation}
\frac{(\pi_{t} - \hat{\mu}_{t})^2}{2\hat{\std}_{t}^2} =\frac{(\pi_{t} - \logit_{t})^2}{2\std_t^2} + \frac{(\pi_{t} - \hat{\mu}_{t-1})^2}{2\hat{\std}_{t-1}^2}.
\end{equation}
Therefore, the best estimate of the current liveness $\hat{\mu}_{t}$ and its uncertainty $\hat{\std}_{t}$ can be derived as:
\begin{equation}
\label{Eq.AMA}
\begin{split}
\hat{\mu}_{t} &= \frac{\hat{\std}_{t-1}^2\logit_{t} + \std_{t}^2\hat{\mu}_{t-1}}{\std_{t}^2 + \hat{\std}_{t-1}^2} = \frac{\hat{\std}_{t-1}^2}{\std_{t}^2 + \hat{\std}_{t-1}^2} \logit_{t} + \frac{\std_{t}^2}{\std_{t}^2 + \hat{\std}_{t-1}^2} \hat{\mu}_{t-1}, \\
\hat{\std}_{t} &= \frac{\std_{t}^2 \hat{\std}_{t-1}^2}{\std_{t}^2 + \hat{\std}_{t-1}^2}.
\end{split}
\end{equation}
Since $\hat{\mu}_{t-1}$ can be considered as the best measurement of the current state, we can compute $\std_{t}^2 = (\logit_{t} - \hat{\mu}_{t-1})^2$.
To adapt the probability with the uncertainty estimation, the current moving average $\hat{\mu}_t$ is used as the updated probability and the $\hat{\std}_t$ as the estimated uncertainty. In the end, an activation function such as sigmoid can be applied to normalize the liveness estimation.
The complete proof can be found in the supplementary material.

In practice, we can only keep recent liveness predictions to compute the moving average $\hat{\mu}_{t-1}$ and the standard deviation $\hat{\std}_{t-1}$ as a relaxation.
Then, the whole inference process in the deployment stage is depicted in Alg.~\ref{Alg:Uncertainty}.
Interestingly, if we assume that $\theta = \hat{\std}_{t-1}^2/(\std_{t}^2 + \hat{\std}_{t-1}^2)$ is a constant value and discard the uncertainty, the Eq.~\eqref{Eq.AMA} would degrade to the Exponential Moving Average (EMA) \cite{MA_2019_191111}, which is a common technique used in the finance domain.

{\setlength{\textfloatsep}{0pt}
\begin{algorithm}[t]
     \SetKwInOut{Input}{Input}
     \SetKwInOut{Output}{Output}
     \SetAlgoLined
    \Input{Current video stream $\frm_{t}$, window size $w$}
    \Output{Calibrated probability $\hat{\mu}_{t}$ and estimated uncertainty $\hat{\std}_{t}$}
    Obtain current liveness logit $\logit_{t}$;\\
    Compute $\hat{\mu}_{t-1}$ and $\hat{\std}_{t-1}$ with a window size $w$;\\
    Compute the weight $\theta =\frac{\hat{\std}_{t-1}^2}{\std_{t}^2 + \hat{\std}_{t-1}^2}$;\\
    Compute $\hat{\mu}_{t} = \theta \logit_{t} + (1 - \theta) \hat{\mu}_{t-1}$ and $\hat{\std}_{t} = \theta \cdot \std_{t}^2$;
    \caption{Training-Free Uncertainty Module}
    \label{Alg:Uncertainty}
\end{algorithm}}

\subsection{Implementation}
We implement our solution using the MXNet/Gluon \cite{Chen_2015_181026} framework and perform all the experiments on a Nvidia V100 GPU. 
The implementation details are presented in the supplementary material.
The code will be published to fully reproduce the experimental results.

\section{Experiments}
\label{Sec:Exps}
\subsection{Experiments Settings}
\noindent \textbf{Datasets}:
Several public datasets were used to benchmark the proposed method: 
\begin{enumerate*}[label=(\roman*)]
    \item SiW \cite{Liu_2018_190806} dataset consists of $165$ subjects with $4,620$ videos in total to evaluate the robustness of the model with various poses, data sources, and unknown attacks. 
    \item OULU-NLP \cite{Boulkenafet_2017_190908} dataset contains $4,950$ real and attack videos, recorded using six different phone cameras.
    There are four different attacking types: two printer attacks and two video replay attacks. 
    Four protocols were used to evaluate the performance with variations of environmental conditions, unseen attack instruments, unknown mobile acquisition devices, and unseen presentation attacks.
    \item SiW-M \cite{Liu_2019_190813} dataset consists of 493 subjects with up to 13 types of spoofing attacks. 
    The protocols in this dataset were designed for open scenario evaluation. 
    It adopts a leave-one-out setting, using twelve ``attack'' videos as the training set and the remaining one as the testing set.
\end{enumerate*}

\noindent \textbf{Baselines}: 
Several baselines were implemented:
\begin{enumerate*}[label=(\roman*)]
    \item R50: This model was trained using the ResNet-50 network \cite{He_2016_17161} to extract the features of each frame. It served as a baseline of the frame-based face liveness detection system.
    \item R50-LSTM: This model was trained using LSTM to learn the temporal information with the features generated by R50. 
    \item STASN \cite{Yang_2019_190813}: This model explored both spatial and temporal information to make a final prediction. The original paper only reported ACER, so we implemented our own version (denoted as STASN*), which obtained slightly better performance than the original work.
\end{enumerate*}

\noindent \textbf{Model Variants}: 
\acron was developed on R50 network by default. Another lightweight extension using MobileNetv2 \cite{Sandler_2018_200302} backbone with a growth rate of 0.5 was also developed, denoted by \acron-MN, to show the potential of low-power deployment like edge devices.

\noindent \textbf{Evaluation Metrics}:
In addition to some widely used metrics (\eg, APCER, BPCER, ACER, and ROC) suggested by ISO \cite{ISO_FDIS_30107_2017} and Zhang \etal \cite{Zhang_2019_190806} that measures on the frame level, we proposed to report the evaluation metrics based on the video segment level.
The videos were divided by the sequence of video segment with length $K$, and each video segment is treated independently, where the previous mentioned metrics (\eg, ACER and ROC) can be applied. For 30 FPS video, we suggest the maximum latency $K$ of less than 30 (1 second) to have a good user experience.
Such evaluation approach have the following advantages:
\begin{enumerate*}[label=(\roman*)]
    \item Compare to frame-based evaluation, it provides two-dimensional metrics, which fitt better to the practical scenario that cares more about the performance with a specific latency (video segment length);
    \item It allows the use of temporal information to some extent, which could provide a fair comparison with video segment-based models (\eg, 3D Convolutions) in the future.
\end{enumerate*}

For all the experiments, to fairly compare with the previous methods and the baseline models, we strictly followed the evaluation protocols provided in each dataset.
In the cross-domain experiments, in addition to the frame-based metrics, we also compared the model performance with state-of-the-art on the video-segment level.


\begin{table}[tb]
\begin{center}
    \begin{minipage}[t]{0.43\linewidth}
        \caption{Comparison of different model implementations.}
        \label{Tab:AB-Module}
        \begin{adjustbox}{max width=\linewidth}
            \begin{tabular}{l|ccc|c}
            \toprule
            Method & Spatial & Temporal & Uncertainty & ACER (\%)\\
            \midrule
            R50 & \checkmark & & & 0.2849 \\
            R50-SMA &  \checkmark & \checkmark & & 0.0927\\
            R50-LSTM & \checkmark & \checkmark & & 0.0794\\
            \midrule
            \acron-NA & \checkmark & \checkmark & & 0.0632 \\
            \acron-EMA & \checkmark & \checkmark & & 0.0028 \\
            \acron & \checkmark & \checkmark & \checkmark & 0.0003\\
            \bottomrule
            \end{tabular}
        \end{adjustbox}
    \end{minipage}
    \quad
    \begin{minipage}[t]{0.53\linewidth}
        \caption{Ablation study on the hyper-parameters of losses.}
        \label{Tab:AB-Loss}
        \begin{adjustbox}{max width=\linewidth}
            \begin{tabular}{c|c|ccc|cccc}
                \toprule
                Hp & Binary & \multicolumn{7}{c}{Multi-class} \\
                \midrule
                $\beta$ & - & 0 & 1 & 0 & 1 & 1 & 1 & 5 \\
                $\gamma$ & -  & 0 & 0 & 0.5& 0.1& 0.5 & 1.0 & 0.1\\
                \midrule
                ACER (\%) & 3.3 & 2.5  & 2.1  & 1.7& 1.3 & 0.8 & 2.5 & 1.7\\
                \bottomrule
            \end{tabular}
        \end{adjustbox}
    \end{minipage}
\end{center}
\vspace{-2em}
\end{table}

\subsection{Ablation Study}
\noindent \textbf{Module}:
The SiW dataset with protocol one was used to evaluate the effectiveness of various components in the proposed framework.
To fully evaluate the components, different model implementations were configured as follows:
\begin{enumerate*}[label=(\roman*)]
    \item R50-SMA: a simple moving average with a window size of five was used to smooth the R50's predictions;
    \item \acron-NA: a R50 model trained with the proposed temporal consistency loss functions;
    \item \acron-EMA: an EMA with a smoothing factor of 0.1 and a window size of five was used to smooth the predictions of the model \acron-NA.
\end{enumerate*}

Table~\ref{Tab:AB-Module} presents the experimental comparisons on the baselines and proposed modules: 
\begin{enumerate*}[label=(\roman*)]
    \item Comparing R50 with R50-SMA, a simple moving average can help smooth the predictions and reduce the ACER by 3 times;
    \item Comparing R50 with R50-LSTM, the temporal information encoded in LSTM does improve the accuracy by 4 times;
    \item Comparing \acron-NA with R50 and R50-LSTM, it can be observed that, rather than using more complex LSTM, the temporal consistency introduced by the proposed loss functions further increases the accuracy of the model;
    \item The uncertainty module does help to improve the robustness of the predictions. Even compared with the EMA, it still achieved much better performance. If comparing the full model with the R50 model, \acron improved ACER by 1000 times approximately.
\end{enumerate*}

\noindent \textbf{Weights of the Losses}:
OULU-NPU dataset with the protocol one was used to evaluate the hyper-parameters to weight the different loss functions. The results are summarized in the Table~\ref{Tab:AB-Loss}:
\begin{enumerate*}[label=(\roman*)]
    \item Compared with the binary classification, the multi-class training ($\beta=0, \gamma=0$) reduced the error rate from 3.3\% to 2.5\%.
    \item To compare the two proposed loss functions, ACER would decrease to $2.1$ if the temporal consistency loss $L_t$ ($\beta=1, \gamma=0$) was applied. By applying the class consistency loss $L_e$, the ACER would further drop to 1.7\%, showing that $L_e$ had a larger impact than $L_t$.
    \item To balance the loss of the three losses (the weight for multi-classification loss is 1), the final ACER could be further reduced from $2.5$ to $0.8$ when $\beta=1$ and $\gamma=0.5$, demonstrating the necessity and effectiveness of both loss functions.
\end{enumerate*}

\noindent \textbf{Pre-trained Models}:
To analyze the impact of using different pre-trained weights as initialization of the backbone R50 model, we tried three different settings: 
\begin{enumerate*}[label=(\roman*)]
    \item Random initialized weights;
    \item Initialization using the pre-trained R50 weights trained on VGGFace-2 dataset. We used cropped face images from the VGGFace-2 dataset \cite{Cao_2018_17830} to train a R50 model for face recognition task, whose weights of the convolutional layers were then used to initialize and fine-tune the new model for the face livness detection task;
    \item Initialization using the pre-trained R50 weights trained on ImageNet \cite{Deng_2009_0214}.
\end{enumerate*}
Evaluation results on the SiW dataset are depicted in Table~\ref{Tab:AB-Pretrain}. 
Surprisingly, the model initialized from the ImageNet dataset achieved a better performance than the model trained from the VGGFace-2 dataset.
One possible reason is that the ImageNet dataset provides a wider distribution of the features, which can better capture the clue of presentation attacks rather than a face recognition dataset.
Though all models are based on the R50 architecture, we believe this conclusion is generalizable to other model architectures as well.

\begin{table}[tb]
    \centering
    \begin{minipage}[t]{.46\linewidth}
        \caption{Comparison on different pre-trained weights as initializations.}
        \label{Tab:AB-Pretrain}
        \begin{adjustbox}{max width=\linewidth}
            \begin{tabular}{lccc}
            \toprule
            Datasets &  APCER (\%) & BPCER (\%) & ACER (\%)\\
            \midrule
            NA & 2.49 & 3.13 & 2.81\\
            VGGFace-2 & 0.76 & 0.64 & 0.70\\
            ImageNet & 0.30 & 0.26 & 0.28\\
            \bottomrule
            \end{tabular}
        \end{adjustbox}
    \end{minipage}
    \hfill
    \begin{minipage}[t]{0.51\linewidth}
        \caption{Runtime comparison with the state-of-the-arts methods.}
        \label{Tab:FLOPs}
        \begin{adjustbox}{max width=\linewidth}
        \begin{tabular}{lcccc}
        \toprule
        Metrics & STASN \cite{Yang_2019_190813} & DTN \cite{Liu_2019_190813} & \acron & \acron-MN \\
        \midrule
        FLOPs (G) & 8* & 6 & 4 & 0.08 \\
        Size (M) & 208 & - & 90 & 2.8\\
        Time (ms) & 11.8 & - & 3.8 & 0.9\\
        \bottomrule
        \end{tabular}
        \end{adjustbox}
    \end{minipage}
    \vspace{-1em}
\end{table}

\subsection{Comparison with the State-of-the-Arts }
\noindent \textbf{Runtime Complexity}:
Inference efficiency is very critical to the low-latency online applications. The total number of floating-point operations (FLOPs) of the model was used to measure the runtime complexity.
 Note that ``*'' denotes the estimated FLOPs based on our implementation because this information cannot be found in the literature.
The lower the FLOPs are, the fewer operations are performed and thus the faster of the inference speed.
As summarized in Table~\ref{Tab:FLOPs}, our model has the least operations, demonstrating better run-time efficiency (\acron only takes 3.8 ms to infer on a single frame, which is approximately three times faster than STASN \cite{Yang_2019_190813}). 
If switching to a lightweight backbone, like \acron-MN, FLOPs and inference time reduced dramatically, showing great potential for low-latency low-power applications.

\begin{figure}[tb]
\begin{center}
    \begin{subfigure}{.55\linewidth}
    \centering
    \includegraphics[width=0.48\linewidth]{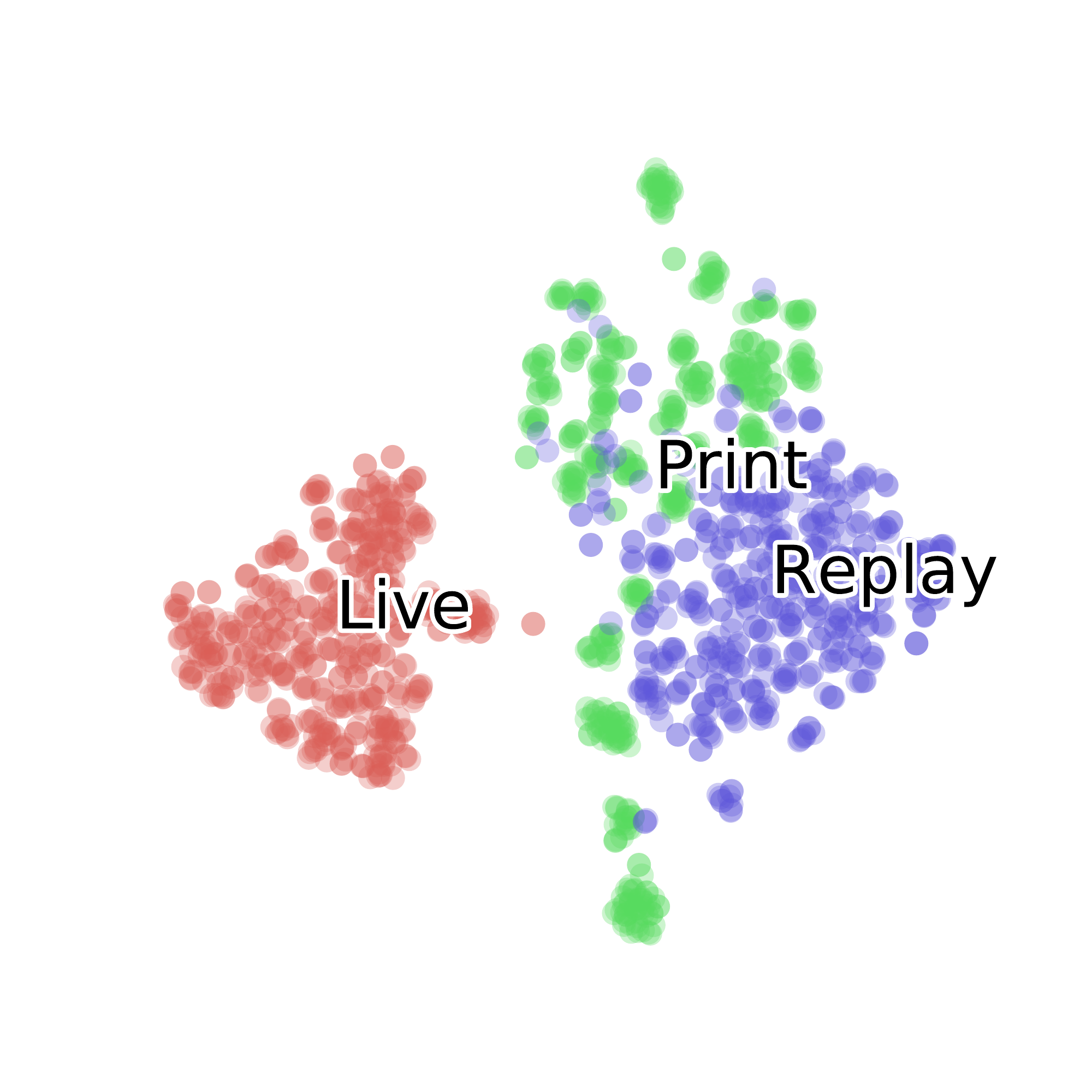}
    \includegraphics[width=0.48\linewidth]{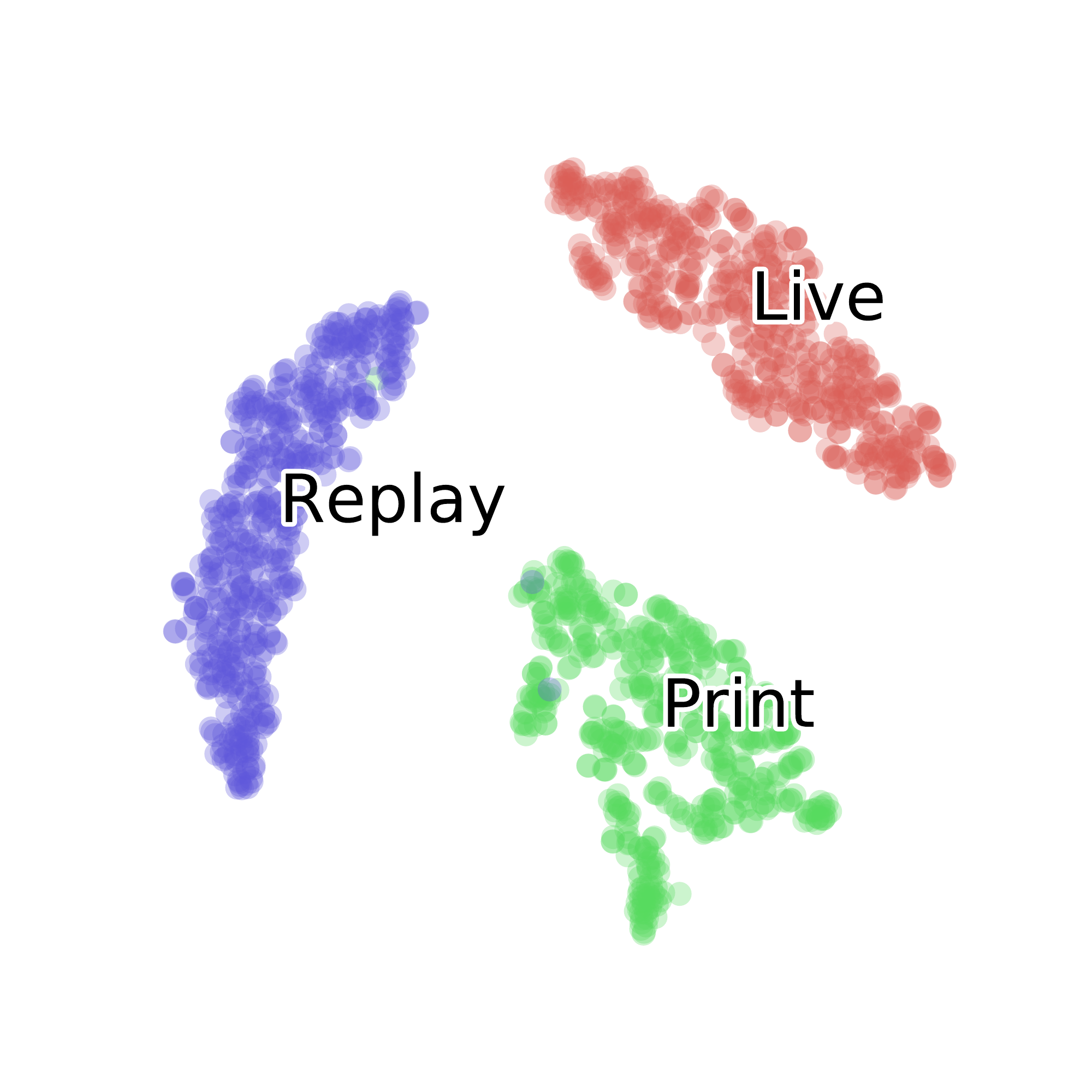}
    \vspace{-0.6em}
    \caption{}
    \label{Fig:SiW-tSNE}
    \end{subfigure}
    \hfill
    \begin{subfigure}{.4\linewidth}
    \centering
    \includegraphics[width=\linewidth]{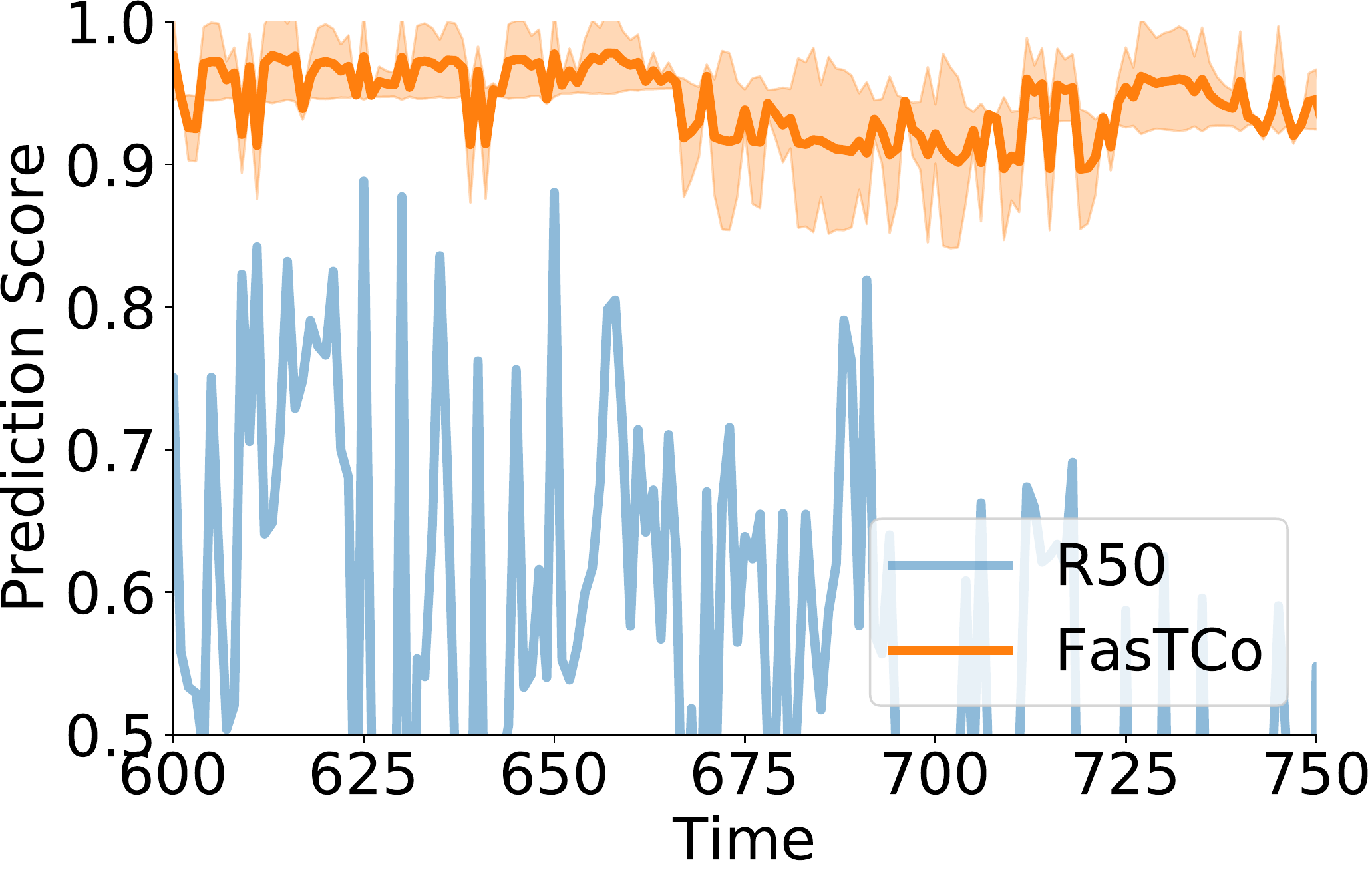}
    \caption{}
    \label{Fig:SiW-TC}
    \end{subfigure}
\end{center}
\vspace{-1.5em}
\caption{Depiction of (a) 2D t-SNE visualization of the representations generated by R50 (Left) and FasTCo (Right) on the SiW dataset; and (b) one sample video liveness predictions across time from R50 and \acron models.}
\vspace{-2em}
\end{figure}

\noindent \textbf{In-the-Wild Scenario}:
The SiW dataset \cite{Liu_2018_190806} was used to evaluate the face liveness detection system in the presence of variances of subject pose, environment illumination, and unseen presentation attacks.
The comparison with the current state-of-the-art methods on this benchmark is summarized in Table~\ref{Tab:SiW}:
\begin{enumerate*}[label=(\roman*)]
    \item The baseline model (R50) initialized with ImageNet pre-trained weights, without any additional data, has already achieved comparable performance to the current state-of-the-art, STASN$^+$, which used additional synthetic augmented data during training.
    One possible reason is that it is easier for the optimizer to find a better local minimum when using this single CNN network rather than training CNN and RNN jointly.
    \item \acron achieved significantly better performance on protocol one, even using the lightweight backbone. One possible explanation is as follows: the subjects in the training set are frontal faces only, leading to slight overfitting to frontal faces for the trained model. However, the subjects in the test videos have more pose changes. Due to the temporal consistency introduced in our uncertainty module, the large variation on the predictions due to the pose change was extremely suppressed, which results in predictions of higher confidence. 
    \item \acron outperformed the state-of-the-art methods by at least 65\% using the last two protocols. Even in protocol three for the open-set scenario, our method achieved better performance, demonstrating the effectiveness of the proposed model for liveness detection in the wild. 
\end{enumerate*}
To visually understand the learned model, the feature representations generated by our baseline R50 and \acron on the testing set are plotted in Fig.~\ref{Fig:SiW-tSNE}.
Compared with the baseline model, the representations of each class produced by our method are more compactly clustered and clearly separated, which could possibly explain the better classification performance.
As a qualitative comparison on the prediction scores with the baseline depicted in Fig.~\ref{Fig:SiW-TC}, our proposed loss functions improved the temporal consistency compared with the baseline while the uncertainty estimation module can further improve the prediction quality during inference.

\begin{table}[tb]
\begin{center}
\caption{Comparisons on the SiW dataset with three protocols (Pr.). The best results are marked in gray.}
\label{Tab:SiW}
\begin{adjustbox}{max width=\linewidth}
\begin{tabular}{c|l|ccc|cc|>{\columncolor[gray]{0.92}}cc}
\toprule
Pr. & Metrics (\%) & TD-SF-CS \cite{Zhang_2019_190806} & STASN* \cite{Yang_2019_190813} & STASN$^{+}$ \cite{Yang_2019_190813} & R50 & R50-LSTM & \acron & \acron-MN\\
\midrule
\multirow{3}{*}{1} & APCER  & 1.27 & 0.72 & - & 0.30 & 0.13 & $0.06 \times 10^{-2}$ & $0.70 \times 10^{-2}$\\ 
& BPCER & 0.33 & 0.89 & - & 0.26 & 0.02 & $0.00 \times 10^{-2}$ & $0.15 \times 10^{-2}$\\ 
& ACER & 0.80 & 0.81 & 0.30 & 0.28 & 0.08 & $0.03 \times 10^{-2}$ & $0.43 \times 10^{-2}$\\
\midrule
\multirow{3}{*}{2} & APCER  & $0.08 \pm 0.17$ & $0.29 \pm 0.16$ & - & $0.08 \pm 0.05$ & $0.03 \pm 0.02$ & 0.02 $\pm$ 0.02 & 0.02 $\pm$ 0.02\\ 
& BPCER & $0.25 \pm 0.22$ & $0.27 \pm 0.14$ & - & $0.07 \pm 0.03$ & $0.03 \pm 0.03$ & 0.00 $\pm$ 0.00  & 0.01 $\pm$ 0.01\\ 
& ACER & $0.17 \pm 0.16$ & $0.28 \pm 0.15$ & $0.15 \pm 0.05$ & $0.03 \pm 0.03$ & $0.03 \pm 0.03$ & 0.01 $\pm$ 0.01 & 0.01 $\pm$ 0.02\\
\midrule
\multirow{3}{*}{3} & APCER  & $6.27 \pm 4.36$ & $11.05 \pm 3.30$ & - & $9.18 \pm 4.32$ & 4.45 $\pm$ 0.51 & 2.73 $\pm$ 0.91 & 3.36 $\pm$ 1.94 \\ 
& BPCER & $6.43 \pm 4.42$ & $7.74 \pm 3.08$ & - & $8.41 \pm 0.94$ & 3.61 $\pm$ 0.67 & 1.28 $\pm$ 0.21 & 5.00 $\pm$ 0.36 \\ 
& ACER & $6.35 \pm 4.39$ & $9.39 \pm 3.19$ & $5.85 \pm 0.85$ & $8.80 \pm 2.62$ & 4.03 $\pm$ 0.08 & 2.00 $\pm$ 0.56 & 4.18 $\pm$ 1.15 \\
\bottomrule
\end{tabular}
\end{adjustbox}
\end{center}
\vspace{-2em}
\end{table}

\begin{table}[tb]
\begin{center}
\caption{Comparison on the OULU-NPU dataset with four protocols (Pr.). The best performance is marked in gray.}
\label{Tab:OULU}
\begin{adjustbox}{max width=\linewidth}
\begin{tabular}{c|l|ccc}
\toprule
Pr. & Method & APCER (\%) & BPCER (\%) & ACER (\%) \\
\midrule
\multirow{7}{*}{1}  & Auxiliary \cite{Liu_2018_190806} & 1.6 & 1.6 & 1.6\\
  & De-Spoofing \cite{Jourabloo_2018_190910} & 1.2 & 1.7 & 1.5\\
  & STASN$^{+}$ \cite{Yang_2019_190813} & 1.2 & 0.8 & 1.0 \\
\rowcolor{lightgray!40} \cellcolor{white} & CDCN++ \cite{Yu_2020_200430} & 0.4 & 0.0 & 0.2 \\
\cmidrule(lr){2-5} 
  & R50 & 2.3 & 4.7 & 3.5\\
  & R50-LSTM & 3.3 & 0.8 & 2.1\\
 & \acron &  0.8 & 0.8 & 0.8\\
\midrule
\multirow{7}{*}{2}  & Auxiliary \cite{Liu_2018_190806} & 2.7 & 2.7 & 2.7\\
  & De-Spoofing \cite{Jourabloo_2018_190910} & 4.2 & 4.4 & 4.3\\
  & STASN$^{+}$ \cite{Yang_2019_190813} & 1.4 & 0.8 & 1.1 \\
  & CDCN++ \cite{Yu_2020_200430} & 1.8 & 0.8 & 1.3 \\
  \cmidrule(lr){2-5}
  & R50 &  2.0 & 1.1 & 1.6\\
  & R50-LSTM & 3.4 & 1.3 & 2.3\\
 \rowcolor{lightgray!40} \cellcolor{white} & \acron &  1.0 & 1.3 & 1.1\\
\bottomrule
\end{tabular}
\quad
\begin{tabular}{c|l|ccc}
\toprule
Pr. & Method & APCER (\%) & BPCER (\%) & ACER (\%) \\
\midrule
\multirow{7}{*}{3}  & Auxiliary \cite{Liu_2018_190806} & 2.7 $\pm$ 1.3 & 3.1 $\pm$ 1.7 & 2.9 $\pm$ 1.5\\
  & De-Spoofing \cite{Jourabloo_2018_190910} & 4.0 $\pm$ 1.8 & 3.8 $\pm$ 1.2 & 3.6 $\pm$ 1.5\\
& STASN$^{+}$ \cite{Yang_2019_190813} & 1.4 $\pm$ 1.4 & 3.6 $\pm$ 4.6 & 2.5 $\pm$ 2.2 \\ 
& CDCN++ \cite{Yu_2020_200430} & 1.7 $\pm$ 1.5 & 2.0 $\pm$ 1.2 & 1.8 $\pm$ 0.7 \\
\cmidrule(lr){2-5}
& R50 & 3.4 $\pm$ 3.0 & 0.7 $\pm$ 1.0 & 2.0 $\pm$ 1.9\\
& R50-LSTM & 4.7 $\pm$ 1.4 & 2.6 $\pm$ 4.2 & 3.7 $\pm$ 2.7\\
\rowcolor{lightgray!40}  \cellcolor{white}  & \acron &  1.2 $\pm$ 1.3 & 1.0 $\pm$ 1.0 & 1.1 $\pm$ 0.8\\
\midrule
\multirow{7}{*}{4}  & Auxiliary \cite{Liu_2018_190806} & 9.3 $\pm$ 5.6 & 10.4 $\pm$ 6.0 & 9.5 $\pm$ 6.0\\
& De-Spoofing \cite{Jourabloo_2018_190910} & 5.1 $\pm$ 6.3 & 6.1 $\pm$ 5.1 & 5.6 $\pm$ 5.7\\
& STASN$^{+}$ \cite{Yang_2019_190813} & 0.9 $\pm$ 1.8 & 4.2 $\pm$ 5.3 & 2.6 $\pm$ 2.8\\ 
& CDCN++ \cite{Yu_2020_200430} & 4.2 $\pm$ 3.4 & 5.8 $\pm$ 4.9 & 5.0 $\pm$ 2.9 \\
\cmidrule(lr){2-5}
& R50 & 5.1 $\pm$ 3.9 & 4.1 $\pm$ 2.4 & 4.6 $\pm$ 2.1\\
& R50-LSTM & 8.9 $\pm$ 7.6 & 4.6 $\pm$ 3.7 & 6.7 $\pm$ 3.6 \\
\rowcolor{lightgray!40}   \cellcolor{white}  & \acron &  1.0 $\pm$ 2.0 & 2.0 $\pm$ 4.1 & 1.5 $\pm$ 1.2\\
\bottomrule
\end{tabular}
\end{adjustbox}
\end{center}
\vspace{-2em}
\end{table}

\noindent \textbf{Mobile Scenario}:
The comparison with the state-of-the-art on the OULU-NPU dataset \cite{Boulkenafet_2017_190908} is depicted in Table~\ref{Tab:OULU}.
Similarly, our single network method outperformed the state-of-the-art on this benchmark on three out of four protocols.
Note that STASN$^{+}$ consisted of multiple networks (R50+LSTM for extracting temporal information and R50 for local spatial information). Besides it was trained with additional synthetic data, while \acron only used the provided training set. 
However, it obtained comparable performance using protocol one and two and achieved at least 40\% improvement using protocol three and four, indicating more robustness to acquisition device changes, unseen illumination conditions, and unseen presentation attacks.

\begin{table}[tb]
\begin{center}
\caption{Comparison on SiW-M dataset with open-set evaluation protocols. The best overall performance is marked in gray.}
\label{Tab:SiW-M}
\begin{adjustbox}{max width=\linewidth}
\begin{tabular}{llcccccccccccccc}
\toprule
\multirow{2}{*}{Method} & \multirow{2}{*}{Metrics (\%)} & \multirow{2}{*}{Replay} & \multirow{2}{*}{Print} & \multicolumn{5}{c}{Mask Attacks} & \multicolumn{3}{c}{Makeup Attacks} & \multicolumn{3}{c}{Partial Attacks} & \multirow{2}{*}{Average} \\ \cmidrule(lr){5-9} \cmidrule(lr){10-12} \cmidrule(lr){13-15}
&&&& Half & Sili. & Trans. & Paper & Manne. & Obf. & Imp. & Cos. & F. Eye & P. Glass & P. Paper & \\
\midrule
 & APCER & 23.7& 7.3 & 27.7 & 18.2 & 97.8 & 8.3 & 16.2 & 100.0 & 18.0 & 16.3 & 91.8 & 72.2 & 0.4 & 38.3 $\pm$ 37.4\\
\rowcolor{lightgray!40} \cellcolor{white} & BPCER & 10.1 & 6.5 & 10.9 & 11.6 & 6.2 & 7.8 & 9.3 & 11.6 & 9.3 & 7.1 & 6.2 & 8.8 & 10.3 & 8.9 $\pm$ 2.0 \\
 & ACER  & 16.8 & 6.9 & 19.3 & 14.9 & 52.1 & 8.0 & 12.8 & 55.8 & 13.7& 11.7& 49.0& 40.5& 5.3 & 23.6 $\pm$ 18.5\\
\multirow{-4}{*}{Auxiliary \cite{Liu_2018_190806}} & EER &14.0 & 4.3 & 11.6 & 12.4 & 24.6& 7.8 & 10.0& 72.3 & 10.1 & 9.4& 21.4& 18.6 & 4.0 & 17.0 $\pm$ 17.7\\
\midrule
\multirow{4}{*}{DTN \cite{Liu_2019_190813}} & APCER & 1.0& 0.0 & 0.7 & 24.5 & 58.6 & 0.5 & 3.8 & 73.2 & 13.2 & 12.4 & 17.0 & 17.0 & 0.2 & 17.1 $\pm$ 23.2\\
 & BPCER & 18.6 & 11.9 & 29.3 & 12.8 & 13.4 & 8.5 & 23.0 & 11.5 & 9.6 & 16.0& 21.5 & 22.6 & 16.8 & 16.6 $\pm$ 6.2 \\
 & ACER  & 9.8 & 6.0 & 15.0 & 18.7 & 36.0 & 4.5 & 7.7 & 48.1 & 11.4& 14.2& 19.3& 19.8& 8.5 & 16.8 $\pm$ 11.1\\
 & EER &10.0 & 2.1 & 14.4 & 18.6 & 26.5& 5.7 & 9.6& 50.2& 10.1& 13.2& 19.8& 20.5& 8.8 & 16.1 $\pm$ 12.2\\
\midrule
\multirow{4}{*}{R50} & APCER &33.9& 2.5 & 17.8 & 9.1 & 20.4 & 0.0 & 0.0 & 52.5 & 0.0 & 13.6 & 58.1 & 0.0 & 12.8 & 17.0 $\pm$ 19.0\\
& BPCER & 7.0 & 6.9 & 15.9 & 9.6 & 1.1 &12.4 & 31.6 & 3.4 & 22.1 &17.1 & 5.9 & 11.0 & 18.9 &  12.5 $\pm$ 8.1\\
& ACER  & 20.5 & 4.7 & 16.8 & 9.3 & 10.8 & 6.2 & 15.8 & 27.9 & 11.0 & 15.4 & 32.0& 5.5& 15.9 & 14.8 $\pm$ 8.0\\
 & EER & 3.0 & 18.0 & 17.0 & 9.0 & 2.0 & 2.0 & 16.0 & 17.0 & 0.0 & 14.0 & 21.0 & 0.0 & 18.0 & 11.0 $\pm$ 8.0\\
\midrule
\rowcolor{lightgray!40} \cellcolor{white} & APCER & 1.7& 0.0 & 2.8 & 9.9 & 3.4 & 0.0& 0.0& 22.9& 0.0& 10.1& 40.9& 12.0 & 0.0& 8.0 $\pm$ 12.0\\
 & BPCER & 20.6 & 12.0 & 10.2 & 9.4 &  20.7 & 9.2 & 14.6 & 10.8 & 8.1 & 9.9 & 8.5& 9.5 & 14.8 & 12.2 $\pm$ 4.3 \\
 \rowcolor{lightgray!40} \cellcolor{white} & ACER  & 11.2 & 6.0 & 6.5 & 9.7 & 12.1& 4.6& 7.3 & 16.8& 4.0& 10.0& 24.7& 10.7& 7.4& 10.1 $\pm$ 5.6\\
 \rowcolor{lightgray!40} \cellcolor{white} \multirow{-4}{*}{\acron} & EER & 7.1 & 5.6 & 7.7 & 9.8 & 14.8& 0.0 & 2.3 & 14.0 & 1.0 & 10.0 & 14.9 & 10.3 &1.8  & 7.6 $\pm$ 5.3\\
\bottomrule
\end{tabular}
\end{adjustbox}
\end{center}
\vspace{-2em}
\end{table}

\begin{figure}[tb]
\begin{center}
    \begin{minipage}{0.46\linewidth}
        \centering
        \includegraphics[width=0.85\linewidth]{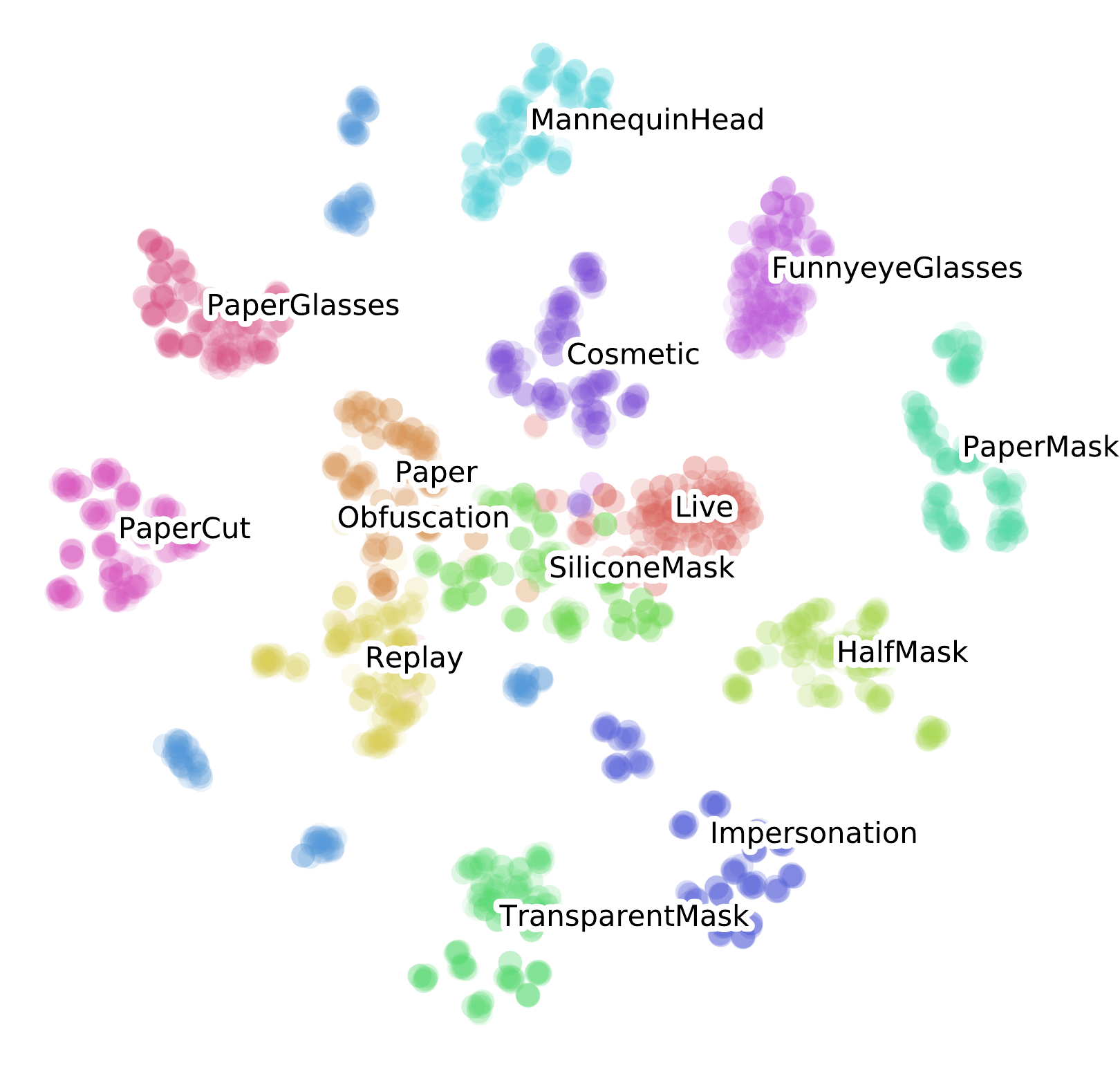}
        \vspace{-1em}
        \caption{2D t-SNE Visualization of the representations of \acron in the open-set liveness detection scenario (Best view in color and zoom in).}
        \label{Fig:IN-TEST-SiW-M-tSNE}
    \end{minipage}
    ~
    \begin{minipage}{0.51\linewidth}
        \centering
        \includegraphics[width=\linewidth]{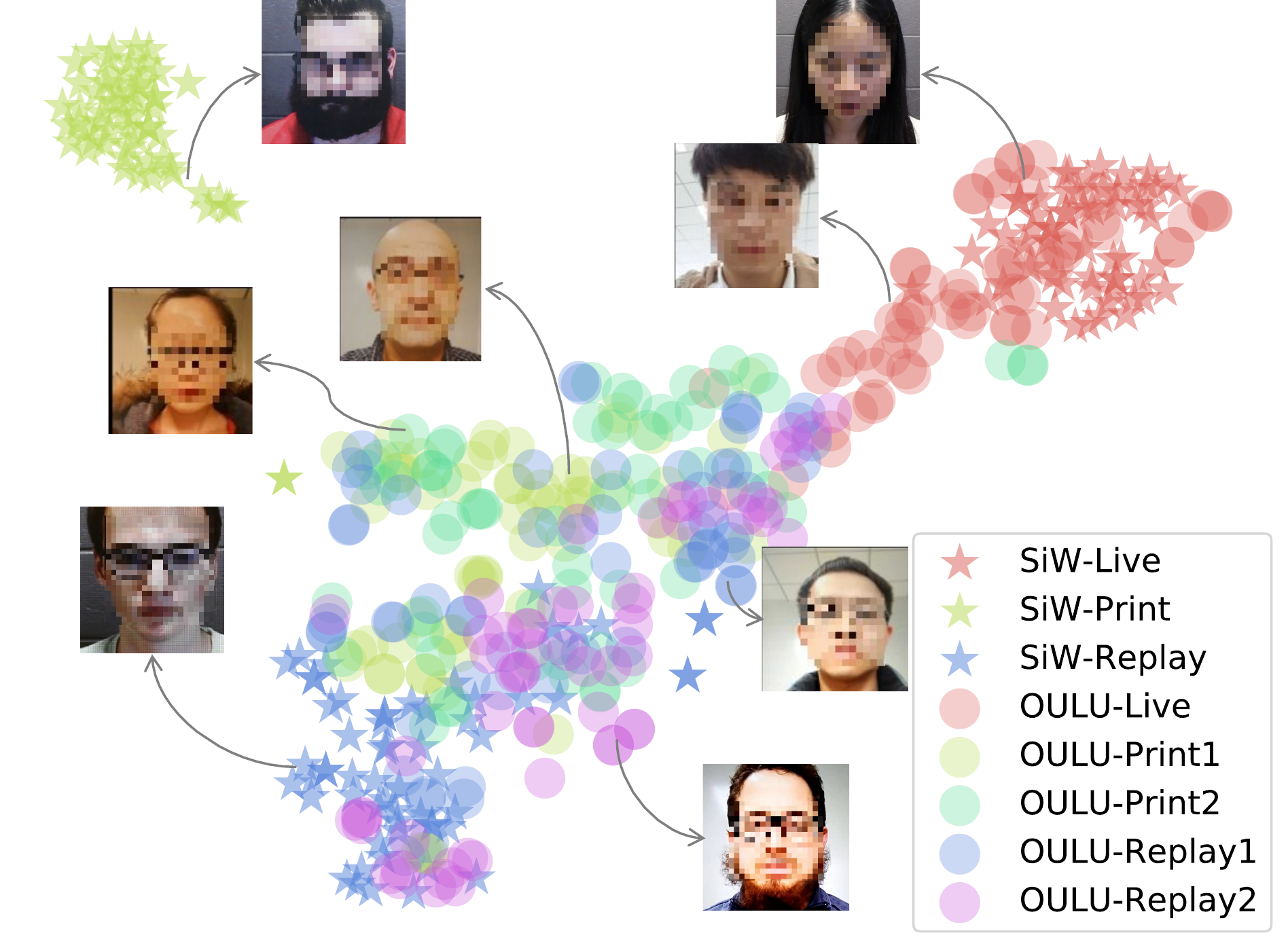}
        \vspace{-2.1em}
        \caption{2D t-SNE visualization of the feature representations generated from SiW and OULU-NPU datasets by \acron trained on SiW dataset only.}
        \label{Fig:SiW-OULU-tSNE}
    \end{minipage}
\end{center}
\vspace{-2.5em}
\end{figure}

\begin{table}[tb]
\begin{center}
\caption{Comparison with the state-of-the-art in a cross-domain setting (Best results are marked in gray).}
\label{Tab:SiW-OULU}
\begin{adjustbox}{max width=\linewidth}
\begin{tabular}{l|ccc|c|>{\centering\arraybackslash}p{1.2cm}>{\centering\arraybackslash}p{1.2cm}>{\centering\arraybackslash}p{1.2cm}>{\centering\arraybackslash}p{1.2cm}>{\centering\arraybackslash}p{1.2cm}}
\toprule
\multirow{2}{*}{Method} & \multirow{2}{*}{APCER (\%)} & \multirow{2}{*}{BPCER (\%)} & \multirow{2}{*}{ACER (\%)} & \multirow{2}{*}{EER (\%)} & \multicolumn{5}{c}{FNR(\%)@FPR=}\\ \cmidrule(lr){6-10}
& & & & & 1E-5& 1E-4& 1E-3 & 1E-2 & 1E-1\\ 
\midrule
Auxiliary \cite{Liu_2018_190806} & 26.82& 14.17 & 20.50& 17.07 & 98.07 & 96.91 & 94.70 & 79.26 & 34.45\\
STASN* \cite{Yang_2019_190813} & 13.24& 5.47 & 9.35 & 5.42 & 67.13 & 62.46 & 52.22 & 20.67 & 3.13\\
R50 & 8.32 & 4.48 & 6.40 & 4.28 & 84.38 & 81.93 & 77.15 & 15.60 & 1.64\\
\midrule
\rowcolor{lightgray!40} \acron &  5.14 & 2.44 & 3.79 & 2.57 & 49.69 & 46.25 & 44.68 & 12.94 & 0.10\\
\bottomrule
\end{tabular}
\end{adjustbox}
\end{center}
\vspace{-2em}
\end{table}

\noindent \textbf{Open-world Scenario}: 
The SiW-M dataset \cite{Liu_2019_190813} was used to evaluate the performance of the model when it encounters unseen presentation attacks in the open-world scenario.
Two state-of-the-art methods \cite{Liu_2018_190806,Liu_2019_190813} were reported on this dataset.
Due to the lack of validation set to choose a threshold in this zero-shot scenario, a high threshold of 0.99 was set to reduce the false positive alarms.
The comparison with the recent methods, depicted in Table~\ref{Tab:SiW-M}, demonstrated that our method outperformed the previous methods by at least 40\% in terms of APCER, ACER, and EER.
Diving deep into the details of the unseen attack scenarios, the following observations can be summarized:
\begin{enumerate*}[label=(\roman*)]
    \item In general, our method performed well on detecting paper mask, mannequin head, impersonation, and partial paper or paper cut attacks.
    \item Compared with DTN \cite{Liu_2019_190813}, \acron obtained a worse result on predicting video replay attacks and achieved comparable performance on detecting print attacks. 
    \item Compared with \mbox{DTN \cite{Liu_2019_190813}}, a significant improvement was achieved by \acron in detecting various masks, makeup (especially obfuscation attack), and partial occlusion attacks. It reveals that our uncertainty estimation module using temporal consistency also works well on detecting most of the unseen presentation attacks. 
\end{enumerate*}

To visually understand the performance in seen and unseen attack scenarios, we separated out the video with silicone attacks, a hard case to 2D face liveness detection system, as the unseen presentation attacks. 
Then, a randomly selected 80\% of the other videos were selected as the training set and the rest was used as the seen attacks.
We retrained the model and depicted the representations generated from the testing set in Fig.~\ref{Fig:IN-TEST-SiW-M-tSNE}:
\begin{enumerate*}[label=(\roman*)]
    \item Most samples belong to the same attack types are clustered, even for the unseen silicone mask attack samples. It demonstrates the generalization of our model to this unseen attacks.
    \item There is a small overlap between silicone mask samples and live samples, which explains why this attack is more difficult to detect.
    \item The features from obfuscation are more scattered, indicating that the network suffers from learning a unique representation for this attack.
\end{enumerate*}

\noindent \textbf{Cross-domain Scenario}:
To verify the generalization of the model, the following experiment was designed: 
Because both the SiW and OULU-NPU datasets contain the print and video attacks yet there is a large domain gap between these two, the SiW dataset was selected as the training set and the OULU-NPU dataset was used as the testing set.
The state-of-the-art models such as Auxiliary \cite{Liu_2018_190806} and STASN \cite{Yang_2019_190813} and R50 model were selected as the baselines in this experiment. 
The model of Auxiliary generously provided by the authors was directly used to test its performance.
Table~\ref{Tab:SiW-OULU} shows that \acron achieved significantly better performance (40\%+ in ACER) compared with all baselines across all metrics.
The video-segment level evaluation was performed and the results are depicted in Table~\ref{Tab:SiW-OULU-SEG}.
\acron consistently outperformed all baselines across different video segment lengths.
Considering both model performance and latency, the video segment length between 5 to 15 were highly recommended in practice.
Figure~\ref{Fig:SiW-OULU-tSNE} depicts the t-SNE visualization of feature representations extracted from two different domains. 
\begin{enumerate*}[label=(\roman*)]
    \item The model correctly learned the liveness features directly from videos since the features from two different domains were highly clustered.
    \item The feature representations from SiW-Print were isolated to the other features while the features from SiW-Replay were cluttered with the attack features from the OULU-NPU dataset, which indicates that the clue to distinguish the presentation attacks from the OULU-NPU dataset was mostly learned from Replay attacks in the SiW dataset.
\end{enumerate*}
In summary, \acron has better generalization for cross-domain applications than current state-of-the-arts, and the actual performance will be even better if the target domain has higher overlap with the source domain. 

\begin{table}[tb]
\begin{center}
\caption{Comparison with the state-of-the-art in a cross-domain setting using the proposed segment level evaluation (Best results are marked in gray).}
\label{Tab:SiW-OULU-SEG}
\begin{adjustbox}{max width=\linewidth}
\begin{tabular}{p{2cm}>{\centering\arraybackslash}p{3cm}>{\centering\arraybackslash}p{1.2cm}>{\centering\arraybackslash}p{1.2cm}>{\centering\arraybackslash}p{1.2cm}>{\centering\arraybackslash}p{1.2cm}>{\centering\arraybackslash}p{1.2cm}>{\centering\arraybackslash}p{1.2cm}}
\toprule
\multirow{2}{*}{Method} & \multirow{2}{*}{Metrics (\%)} & \multicolumn{6}{c}{video segment length $K=$}\\ \cmidrule(lr){3-8}
& & 1 & 3& 5 & 10 & 15 & 30 \\ 
\midrule
\multirow{2}{*}{STASN* \cite{Yang_2019_190813}} & ACER & 9.35 & 9.11 & 8.87 & 8.74 & 9.17 & 8.91\\
& FNR@FPR=1E-2 & 20.67 & 20.86 & 19.50 & 17.53 & 18.77 & 18.08\\
\midrule
\multirow{2}{*}{R50} & ACER & 6.40 & 6.08 & 5.98 & 5.87 & 5.61 & 5.61\\
& FNR@FPR=1E-2 & 15.60 & 16.70 & 16.65& 15.60 & 16.21& 14.15\\
\midrule
\rowcolor{lightgray!40}  &  ACER & 3.96 & 3.92 & 3.87 & 3.77 & 3.70 & 3.69\\
\rowcolor{lightgray!40} \multirow{-2}{*}{\acron} & FNR@FPR=1E-2 & 9.72 & 8.94 & 8.55 & 8.60 & 8.28 & 7.86\\
\bottomrule
\end{tabular}
\end{adjustbox}
\end{center}
\vspace{-2em}
\end{table}



\section{Conclusion}
\label{Sec:Con}
In this paper, the temporal inconsistency was identified as a common underlying issue that undermines the model performance in the face liveness detection task.
To address this issue, in addition to the classification loss, temporal consistency and class consistency losses were proposed for training the model.
Moreover, a training-free uncertainty estimation module was developed to update the prediction adaptively in a smooth manner.
Extensive experiments have demonstrated that, by applying two proposed strategies based on temporal consistency, the model outperformed the current state-of-the-art by a significant margin.

\bibliographystyle{splncs04}
\bibliography{egbib}

\end{document}